# Tell Me Who Your Students Are: GPT Can Generate Valid Multiple-Choice Questions When Students' (Mis)Understanding Is Hinted

Machi Shimmei[1,5], Masaki Uto[2], Yuichiroh Matsubayashi[1,5],
Kentaro Inui[3,1,5], Aditi Mallavarapu[4] and Noboru Matsuda[4]

[1]Tohoku University, Sendai, Japan
{machi.shimmei.e6, y.m}@tohoku.ac.jp
[2]The University of Electro-Communications, Tokyo, Japan
uto@ai.lab.uec.ac.jp
[3]MBZUAI, Abu Dhabi, United Arab Emirates
kentaro.inui@mbzuai.ac.ae
[4]North Carolina State University, Raleigh, USA
{amallav, noboru.matsuda}@ncsu.edu
[5]RIKEN, Wako, Japan

**Abstract.** The primary goal of this study is to develop and evaluate an innovative prompting technique, AnaQuest, for generating multiple-choice questions (MCQs) using a pre-trained large language model. In AnaQuest, the choice items are sentence-level assertions about complex concepts. The technique integrates formative and summative assessments. In the formative phase, students answer open-ended questions for target concepts in free text. For summative assessment, AnaQuest analyzes these responses to generate both correct and incorrect assertions. To evaluate the validity of the generated MCQs, Item Response Theory (IRT) was applied to compare item characteristics between MCQs generated by AnaQuest, a baseline ChatGPT prompt, and human-crafted items. An empirical study found that expert instructors rated MCQs generated by both AI models to be as valid as those created by human instructors. However, IRT-based analysis revealed that AnaQuest-generated questions—particularly those with incorrect assertions (foils)—more closely resembled human-crafted items in terms of difficulty and discrimination than those produced by ChatGPT.

## 1 Introduction

Multiple-choice questions (MCQ) are among the most widely used types in authentic educational settings [1-3]. While MCQs take many forms, this study focuses on *analytical MCQs* on a complex topic with a stem: "Which of the following are correct about [topic]?". Their answer choices consist of *assertions* that require analytical reasoning rather than mere fact recall to be answered correctly. The effectiveness of analytical MCQs depends on the "validity" of answer choices. Validity can be viewed from multiple perspectives, one of which is ensuring that the choices reflect students' understanding and misconceptions of the topics being assessed. Foils (or "distractors") should strike a balance by being challenging enough to address underlying misconceptions without being too obvious or excessively difficult.

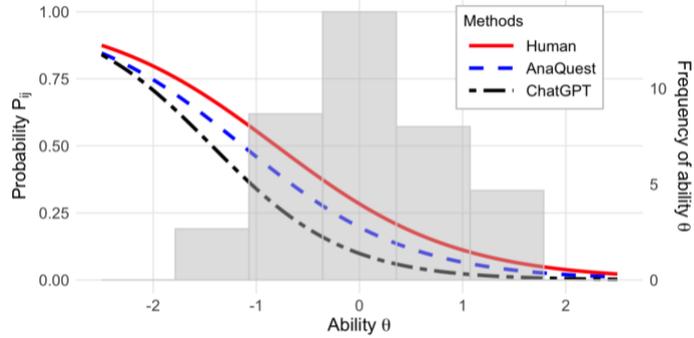

Figure 1: Item characteristic curve of foils generated by different sources (Human, AnaQuest, and ChatGPT). The mean of discrimination and difficulty params estimated by a two-parameter Item-Response Theory model are used. The shaded histogram shows the sample distribution of the estimated student ability parameter. See details in section 5.1.

Generating foils that tap the ability for analytical reasoning requires knowledge about misconceptions that students might develop. We hypothesize that experienced instructors can create valid answer choices based on their observations about how students struggle with learning complex concepts. Nonetheless, creating MCQs is extremely labor-intensive [4, 5], and research has increasingly explored the generative AI, i.e., pre-trained Large Language Models (LLM) to automate it [6, 7]. Accordingly, a critical question arises: Can LLMs make MCQs with validity comparable to those created by experienced instructors?

Figure 1 illustrates how the probability of selecting a foil varies with the student's latent ability. *The plot reveals a gap in the validity of the foils generated by the bare-bones ChatGPT (vanilla adoption of a pre-trained LLM) and those created by an expert instructor, with the latter exhibiting more desirable foil characteristics*. We hypothesize that the ChatGPT's shortfall stems from its pre-training data, which might lack knowledge of students' misconceptions about the topics being assessed, or from the model's failure to retrieve such knowledge during generation.

To address this issue, we propose AnaQuest, a prompting technique for pre-trained LLMs that incorporates students' understanding of a topic and a learning objective into the prompt. Students' understanding is captured through free-text responses to an open-ended question related to the topic, collected as formative assessments. Figure 1 shows that AnaQuest can generate incorrect assertions that are significantly closer to those created by an instructor than those produced by the bare-bones ChatGPT.

Expert surveys evaluating the pedagogical value of AI-generated questions are a commonly used technique [8, 9]. However, our current study demonstrated that such survey-based evaluations might overlook latent psychometric properties, such as discrimination and difficulty, that distinguish AI-generated from human-crafted questions. In this study, we propose to assess the validity of AI-generated MCQs by leveraging proximity matrices (e.g., statistical distance), in combination with Item Response Theory, to evaluate how closely their discrimination and difficulty resemble those of MCQs created by experienced instructors—both at the question and foil levels.

The contributions of this study include the following: (1) We propose AnaQuest, a prompting technique for generating analytical MCQs based on students' understanding on particular learning objectives. To our knowledge, *this is the first approach to use knowledge from formative assessment to generate valid summative assessments*. (2) We evaluated the validity of AI-generated questions both at the question and foil levels by measuring their proximity to human-crafted questions using the Item-Response Theory (IRT) technique on authentic exam data. To our knowledge, *this is the first study to assess the validity of AI-generated MCQs—including both questions and foils—using IRT as a **proximity measure***.

## 2    Related Work

There has been growing interest in applications of pre-trained LLMs for educational question generation [10-14]. Generating MCQs is one of the most intensively studied topics where how to generate foils (aka distractors) is an essential problem [15, 16]. Commonly used techniques linguistically perturbate correct choices [17-19]. Other researchers utilize ontological knowledge to generate foils [20-22]. While those techniques are mostly for term-level distractors, we focus on sentence-level assertions.

Van Campenhout *et al.* [23] evaluated the difficulty, engagement, and persistence of AI-generated questions using student response data. Other studies have employed IRT to estimate the difficulty and discrimination parameters of questions [24, 25]. However, because IRT parameters are not absolute metrics, drawing objective interpretations can be challenging. Our proposed use of IRT-based proximity matrices aims to address this limitation.

## 3    Analytical Multiple-Choice Question

A multiple-choice question (MCQ) consists of a *stem* and a set of correct and incorrect *choice items*. Incorrect choice items are called distractors or foils. In this paper, we examine a type of MCQs that have sentences as choice items. We particularly focus on MCQs with the following form of stem: "Which of the following are correct about [topic]?" followed by three assertions: A, B, and C, as shown in Figure 2.

A unique aspect of the current study is that assertions are carefully crafted for a specific learning objective at a higher level of Bloom's Taxonomy. The example shown in Figure 2 was made for a following learning objective: "Describe the overall structure of contextual design." If different learning objectives are given, different sets of assertions will be crafted for the same topic (i.e., the question stem). The [topic] is a specific theme to be tested. For example, "the overall structure of contextual design" is one ***topic***, and "the meaning of context in the contextual design" is another, though both are about the same ***theme***: "contextual design."

| | |
|---|---|
| **Q**: | Which of the following are correct about the overall structure of contextual design? |
| A: | Contextual design emphasizes understanding user needs in their real environments by interviewing users, modeling their interactions, and iterating design solutions. |
| B: | The contextual design is participatory method where the primary goal is to solicit ideas about a good interface from users by asking about the deficit of the current design and what users would like to see instead. |
| C: | The main focus of contextual design is to develop affinity diagrams, and when the affinity diagram is well developed, other models like cultural or sequence-flow models will not be needed. |

Figure 2: Example of an exam question.

To answer those MCQs correctly, students must analyze and evaluate given information to carry out logical reasoning beyond merely remembering facts. *We hypothesize that when assertions are carefully crafted for learning objectives at a higher level of Bloom's Taxonomy, answering MCQ requires analytical reasoning skills*. We therefore call this type of MCQs the *analytical multiple-choice questions*.

## 4   Research Questions and Methods

We investigate the following specific research questions: (RQ1) How do AI-generated questions hold validity relative to questions created by an expert instructor? (RQ2) How do expert instructors perceive the validity of AI-generate questions? An empirical evaluation study was conducted at a college-level course where MCQs generated by an instructor, AnaQuest, and another baseline generative AI were compared.

### 4.1   AnaQuest: System Implementation

AnaQuest was implemented using OpenAI GPT-4o with few-shot learning [26] with five examples. It was tasked with generating three correct and incorrect assertions for a given topic and a learning objective. In addition, AnaQuest received students' responses to an open-ended question related to the topic. The model was prompted to ensure that the generated assertions could effectively assess whether a student had achieved the learning objective. It was also instructed to generate incorrect assertions that reflect common student misunderstandings identified in the provided responses.

### 4.2   Overview of the Evaluation Study

The evaluation study was conducted using the final exam at a college-level course on Human-Computer Interaction (HCI) in fall 2024. Prior to the exam, students answered two open-ended questions per week for eight weeks—i.e., 16 topics were probed with 50 to 56 student responses per topic. As a result, AnaQuest generated 16 multiple-choice questions (MCQs), each with three correct and three incorrect assertions.

OpenAI ChatGPT (4o Dec. 2024 ver.) [27] was used as a baseline generative AI. 16 questions for the same topics used for AnaQuest was created with a straightforward prompt: "`I need to create a final exam for a graduate level HCI course. Generate three correct and three incorrect choices, each about 30 words, for a multiple choice question asking "Which of the following are correct about [TOPIC]?"`".

An experienced instructor who has been teaching the HCI course for more than 10 years created eight MCQs with slightly modified topics to avoid duplications with the AI-generated questions while the overall themes to be tested remained the same. The instructor also reviewed AI-generated questions and selected three assertions per question to be used for the exam while ensuring their adequacy as exam questions.

Two exam versions were created, each containing all eight human-crafted questions, but a different set of five ChatGPT- and five AnaQuest-generated questions with topic coverage counterbalanced. Each question had 9 answer choices, including all combinations of correct assertions and an "I do not know" option. Students gained 1 point for correct answers, lost 1/7 for incorrect ones, and earned 0 for "I do not know." Students were informed that this scoring policy was used to discourage random guessing.

### 4.3 Analysis

Multiple-choice questions originated from three *sources* (AnaQuest, ChatGPT, and Human) were comparatively analyzed using an IRT-based method and a survey study.

**The IRT-based method** was conducted both at the *question-* and *foil-levels*. For the question-level analysis, we applied the two-parameter logistic (2PL) model [28, 29]. Questions whose difficulty parameter fell outside the range of the mean ability parameter ± 3 SD were excluded as outliers.

For the foil-level analysis, *incorrect assertions* were treated as items. We operationalize the foil validity as the likelihood that students believe a foil is correct, modeled as a function of their latent ability. To this end, we applied the 2PL IRT model to a response matrix indicating which students considered which foils being correct. Discrimination and difficulty parameters for each foil were computed using the latent student ability scores obtained from the question-level analysis. Notably, valid foils are expected to have negative values for both discrimination and difficulty.

**The survey study** was conducted with three expert instructors who regularly teach the HCI course. Participants were asked to rate each exam question based on the following five criteria, using a 5-Likert scale. The source of question (AnaQuest, ChatGPT, or Human) was hidden in the survey: **S1** (*Answerability*): The question is clear enough to answer. **S2** (*Correctness*): The given answers are correct. **S3** (*Adequate Complexity*): The given answers are adequate, i.e., not too easy, not too difficult, not too confusing, etc. **S4** (*Alignment*): The question is suitable for the learning objective. **S5** (*Adoptability*): I would adopt the question to my course.

## 5 Results

### 5.1 Results from the IRT-based methods

There were 57 students who took the exam—29 and 28 for the exam Ver. A and B, respectively. According to the overall correctness of student answers, the two versions were equivalent in their complexity; the ratio of correct responses to the total number of responses was 0.56±0.02 and 0.55±0.02 for version A and B, respectively: $t(55)=0.17, p=0.86$.

**Question-Level Item Analysis.** Following the rule of outliers (section 4.3), 9 questions (2 Human, 4 ChatGPT, and 3 AnaQuest) were excluded from the analysis. Analysis discrimination showed that AnaQuest-generated questions had discrimination much more like human-crafted questions than ChatGPT-generated questions; overall average of discrimination across questions were 0.89±0.25 for AnaQuest vs. 0.85±0.37 for Human vs. 1.07±0.25 for ChatGPT.

To quantitatively compare the proximity of estimated discrimination $\alpha$ and difficulty $\beta$ among three sources, we estimated the bivariate distributions $(\alpha, \beta)$. We then used KL-Divergence (KLD) with Gaussian kernel density estimation [30]. The results showed that human-crafted and AnaQuest-generated questions are notably more similar (KLD = 820.96) than human-crafted and ChatGPT-generated questions (KLD = 1820.01). KLD between AnaQuest- and ChatGPT-generated questions was 1423.45. *This finding implies that AnaQuest can generate questions with more similar IRT characteristics to human-crafted ones than ChatGPT.*

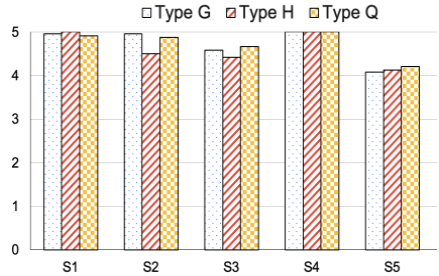

Figure 3: The average ratings per source—Human (H), AnaQuest (Q), and ChatGPT (G)—across participants for each metric.

**Foil-Level Item Analysis.** Figure 1 shows the item characteristic curve (ICC) to visualize the validity of foils. ICC shows the probability that a student $j$ considers a foil $i$ as a correct assertion as a function of the student's ability $\theta_j$. In Figure 1, ICC is overlaid on the distribution of student ability, which is shown as a shaded histogram. The plot shows that AnaQuest-generated foils are more closely approximated human-crafted ones than ChatGPT-generated foils. ChatGPT-generated foils exhibited a notably steeper slope (higher discrimination) and much more left-shifted (lower difficulty). *This suggests that ChatGPT-generated foils were more obviously incorrect and, therefore, more likely to be selected only by very low-ability students.*

To quantify how closely AnaQuest-generated foils resembled human-crafted ones, we again applied the proximity analysis as done in the question-level item analysis. The results showed that the AnaQuest-generated foils had a closer bivariate distribution to Human-crafted ones than ChatGPT; KLD between Human and AnaQuest was 13.66, whereas KLD between Human and ChatGPT was 36.60.

In sum, the foil-level analysis demonstrated that *AnaQuest can produce well-founded foils, fulfilling the critical property of foils for analytical MCQ: low-ability students are more likely to select foils compared to high-ability students.* Furthermore, the analysis revealed that *foils generated by AnaQuest exhibited an appropriate power of discrimination and the level of difficulty, more closely resembling those of human-crafted foils than ChatGPT.*

### 5.2    Results from Instructor Survey

Figure 3 shows the average ratings by source (Human, AnaQuest, and ChatGPT) across three survey participants, based on the five criteria mentioned in section 4.3. Overall, the instructors rated the pedagogical effectiveness similarly across all sources. Notably, for the correctness criterion (S2: The given answers are correct), human-crafted questions scored slightly lower than AI-generated ones. This may be due to ambiguous language in human-written foils, particularly in analytical MCQs, where generative AI may outperform humans in producing clearer assertions.

## 6      Conclusion

We found that a pre-trained LLM can generate valid foils for *analytical multiple-choice questions* regarding a particular topic when a specific learning objective and students' understanding of the topic were given. Our findings suggest that a valid assessment should not only evaluate what the students expected to learn but also identify potential misunderstandings specific to the cohort taking the exam. Our study also demonstrated that an IRT-based analysis can provide a more systematic comparison of AI-generated questions than a survey-based evaluation. These findings highlight critical lessons, emphasizing the need for ethical considerations in evaluation methods as generative AI becomes more widely integrated into authentic learning environments.

This study is limited by the small number of questions analyzed, which constrained the statistical analysis. As IRT requires substantial response data, future work with larger datasets and more instructors is needed to improve generalizability.

**Acknowledgments.** This research reported here was supported by the National Science Foundation Grant No. 2016966 to North Carolina State University. Any opinions, findings, and conclusions expressed in this material are those of the authors and do not necessarily reflect the views of NSF.

## References

1.  Gall, M.D., *The use of questions in teaching.* Review of educational research, 1970. **40**(5): p. 707-721.
2.  Ch, D.R. and S.K. Saha, *Automatic multiple choice question generation from text: A survey.* IEEE Transactions on Learning Technologies, 2018. **13**(1): p. 14-25.
3.  Haladyna, T.M., S.M. Downing, and M.C. Rodriguez, *A review of multiple-choice item-writing guidelines for classroom assessment.* Applied measurement in education, 2002. **15**(3): p. 309-333.
4.  Kurdi, G., et al., *A systematic review of automatic question generation for educational purposes.* International Journal of Artificial Intelligence in Education, 2020. **30**(1): p. 121-204.
5.  Leo, J., et al., *Ontology-based generation of medical, multi-term MCQs.* International Journal of Artificial Intelligence in Education, 2019. **29**: p. 145-188.
6.  Al Faraby, S., A. Adiwijaya, and A. Romadhony, *Review on Neural Question Generation for Education Purposes.* International Journal of Artificial Intelligence in Education, 2024. **34**(3): p. 1008-1045.
7.  Hwang, K., et al. *Towards automated multiple choice question generation and evaluation: aligning with Bloom's taxonomy*. in *International Conference on Artificial Intelligence in Education*. 2024. Springer.
8.  Gorgun, G. and O. Bulut, *Instruction‐Tuned Large‐Language Models for Quality Control in Automatic Item Generation: A Feasibility Study.* Educational Measurement: Issues and Practice, 2024.
9.  Moore, S., et al., *An Automatic Question Usability Evaluation Toolkit*, in *Proceedings of the International Conference on Artificial Intelligence in Education*, A.M. Olney, et al., Editors. 2024, Springer: Switzerland. p. 31-46.

10. Bulathwela, S., H. Muse, and E. Yilmaz. *Scalable educational question generation with pre-trained language models*. in *International Conference on Artificial Intelligence in Education*. 2023. Springer.
11. Elkins, S., et al. *How useful are educational questions generated by large language models?* in *International Conference on Artificial Intelligence in Education*. 2023. Springer.
12. Wang, Z., et al. *Towards human-like educational question generation with large language models*. in *International conference on artificial intelligence in education*. 2022. Springer.
13. Scaria, N., S. Dharani Chenna, and D. Subramani, *Automated Educational Question Generation at Different Bloom's Skill Levels Using Large Language Models: Strategies and Evaluation*, in *Proceedings of the International Conference on Artificial Intelligence in Education*, A.M. Olney, et al., Editors. 2024, Springer: Switzerland. p. 165-179.
14. Lamsiyah, S., et al., *Fine-Tuning a Large Language Model with Reinforcement Learning for Educational Question Generation*, in *Proceedings of the International Conference on Artificial Intelligence in Education*, A.M. Olney, et al., Editors. 2024, Springer: Switzerland. p. 424-438.
15. Dutulescu, A., et al., *Beyond the Obvious Multi-choice Options: Introducing a Toolkit for Distractor Generation Enhanced with NLI Filtering*, in *Proceedings of the International Conference on Artificial Intelligence in Education*, A.M. Olney, et al., Editors. 2024, Springer: Switzerland. p. 242-250.
16. Ch, D.R. and S.K. Saha, *Generation of Multiple-Choice Questions from Textbook Contents of School-Level Subjects.* IEEE Transactions on Learning Technologies, 2023. **16**(1): p. 1-13.
17. Brown, J., G. Frishkoff, and M. Eskenazi. *Automatic question generation for vocabulary assessment*. in *Proceedings of Human Language Technology Conference and Conference on Empirical Methods in Natural Language Processing*. 2005.
18. Correia, R., et al. *Automatic generation of cloze question stems*. in *Computational Processing of the Portuguese Language: 10th International Conference, PROPOR 2012, Coimbra, Portugal, April 17-20, 2012. Proceedings 10*. 2012. Springer.
19. Agarwal, M. and P. Mannem. *Automatic gap-fill question generation from text books*. in *Proceedings of the sixth workshop on innovative use of NLP for building educational applications*. 2011.
20. Ren, S. and K.Q. Zhu. *Knowledge-driven distractor generation for cloze-style multiple choice questions*. in *Proceedings of the AAAI conference on artificial intelligence*. 2021.
21. Faizan, A. and S. Lohmann. *Automatic generation of multiple choice questions from slide content using linked data*. in *Proceedings of the 8th international conference on web intelligence, mining and semantics*. 2018.
22. EV, V. and P.S. Kumar, *Automated generation of assessment tests from domain ontologies.* Semantic Web, 2017. **8**(6): p. 1023-1047.
23. Van Campenhout, R., M. Hubertz, and B.G. Johnson. *Evaluating AI-Generated Questions: A Mixed-Methods Analysis Using Question Data and Student Perceptions*. in *International Conference on Artificial Intelligence in Education*. 2022. Springer.
24. Liu, M., V. Rus, and L. Liu, *Automatic chinese multiple choice question generation using mixed similarity strategy.* IEEE Transactions on Learning Technologies, 2017. **11**(2): p. 193-202.
25. Gherardi, E., et al. *Using Knowledge Graphs to Improve Question Difficulty Estimation from*

*Text*. in *International Conference on Artificial Intelligence in Education*. 2024. Springer.
26. Brown, T., et al., *Language models are few-shot learners.* Advances in neural information processing systems, 2020. **33**: p. 1877-1901.
27. OpenAI, *ChatGPT (4o, Dec 2024 version) [Large language model]*. 2024, https://chat.openai.com/chat.
28. Baker, F.B. and S.-H. Kim, *Item response theory: Parameter estimation techniques*. 2004: CRC press.
29. Linden, W.J. and R.K. Hambleton, *Handbook of modern item response theory. Volume two: Statistical tools*. 2016, CRC Press.
30. Silverman, B.W., *Density Estimation for Statistics and Data Analysis*. 1986, London: Chapman & Hall.